\documentclass{article}

\usepackage[preprint]{neurips_2022}

\usepackage[utf8]{inputenc} 
\usepackage[T1]{fontenc}    
\usepackage{hyperref}       
\usepackage{url}            
\usepackage{booktabs}       
\usepackage{amsfonts}       
\usepackage{nicefrac}       
\usepackage{microtype}      
\usepackage{xcolor}         
\usepackage{amsmath}
\usepackage{graphicx}
\usepackage{algorithm} 
\usepackage{algpseudocode} 
\usepackage{listings}
\usepackage{multirow}
\usepackage{makecell}
\usepackage{natbib}
\setcitestyle{numbers,square}
\title{DropKey}

\author{%
  Bonan Li\thanks{This work was done during Bonan's internship at MT Lab, Meitu} \\
  \And
  Yinhan Hu \\
  \And
  Xuecheng Nie \\
  \And
  Congying Han \\
  \And
  Xiangjian Jiang \\
  \And
  Tiande Guo \\
  \And
  Luoqi Liu \\
}

\begin{document}

\maketitle

\begin{abstract}
  In this paper, we focus on analyzing and improving the dropout technique for self-attention layers of Vision Transformer, which is important while surprisingly ignored by prior works. In particular, we conduct researches on three core questions: First, \emph{what to drop in self-attention layers}? Different from dropping attention weights in literature, we propose to move dropout operations forward ahead of attention matrix calculation and set the Key as the dropout unit, yielding a novel dropout-before-softmax scheme. We theoretically verify that this scheme helps keep both regularization and probability features of attention weights, alleviating the overfittings problem to specific patterns and enhancing the model to globally capture vital information; Second, \emph{how to schedule the drop ratio in consecutive layers}? In contrast to exploit a constant drop ratio for all layers, we present a new decreasing schedule that gradually decreases the drop ratio along the stack of self-attention layers. We experimentally validate the proposed schedule can avoid overfittings in low-level features and missing in high-level semantics, thus improving the robustness and stableness of model training; Third, \emph{whether need to perform structured dropout operation as CNN}? We attempt patch-based block-version of dropout operation and find that this useful trick for CNN is not essential for ViT. Given exploration on the above three questions, we present the novel DropKey method that regards Key as the drop unit and exploits decreasing schedule for drop ratio, improving ViTs in a general way. Comprehensive experiments demonstrate the effectiveness of DropKey for various ViT architectures, \emph{e.g.} T2T and VOLO, as well as for various vision tasks, \emph{e.g.}, image classification, object detection, human-object interaction detection and human body shape recovery.
\end{abstract}

\section{Introduction}

Vision Transformer (ViT)~\cite{vit} has achieved great success for various vision tasks, \emph{e.g.}, image recognition~\cite{t2t, volo, swin,mvit,tnt}, object detection~\cite{detr}, human body shape estimation~\cite{metro}, etc. Prior works mainly focus on researches of patch division, architecture design and task extension. However, the dropout technique for self-attention layer, which plays the essential role to achieve good generalizability, is surprisingly ignored by the community.

Different from the counterpart for Convolutional Neural Networks (CNNs), the dropout in ViT directly utilizes the one in original Transformer designed for Natural Language Processing, which sets attention weights as the  manipulation unit with a constant dropout ratio for all layers. Despite of its simplicity, this vanilla design faces three major problems. First, it breaks the probability distribution of attention weights due to the averaging operation on non-dropout units after softmax normalization. Although this regularizes the attention weights, it still overfits specific patterns locally due to the failure on penalizing score peaks, as shown in Fig.~\ref{fig:motivation} (a) and (b); Second, the vanilla design is sensitive to the constant dropout ratio, since high ratio occurs missing of semantic information in high-level representations while low ratios overfitting in low-level features, resulting in the unstable training process; Third, it ignores the structured characteristic of input patch grid to ViT, which plays an effective role to improve performance with blockwise dropout in CNNs. These three problems degrade the performance and limit the generalizability of ViTs.

Motivated by the above, we propose to analyze and improve the dropout technique in self-attention layer, further pushing forward the frontier of ViTs for vision tasks in a general way. In particular, we focus on three core aspects in this paper:

\begin{figure}[t!]
  \centering
  \includegraphics[width=\textwidth]{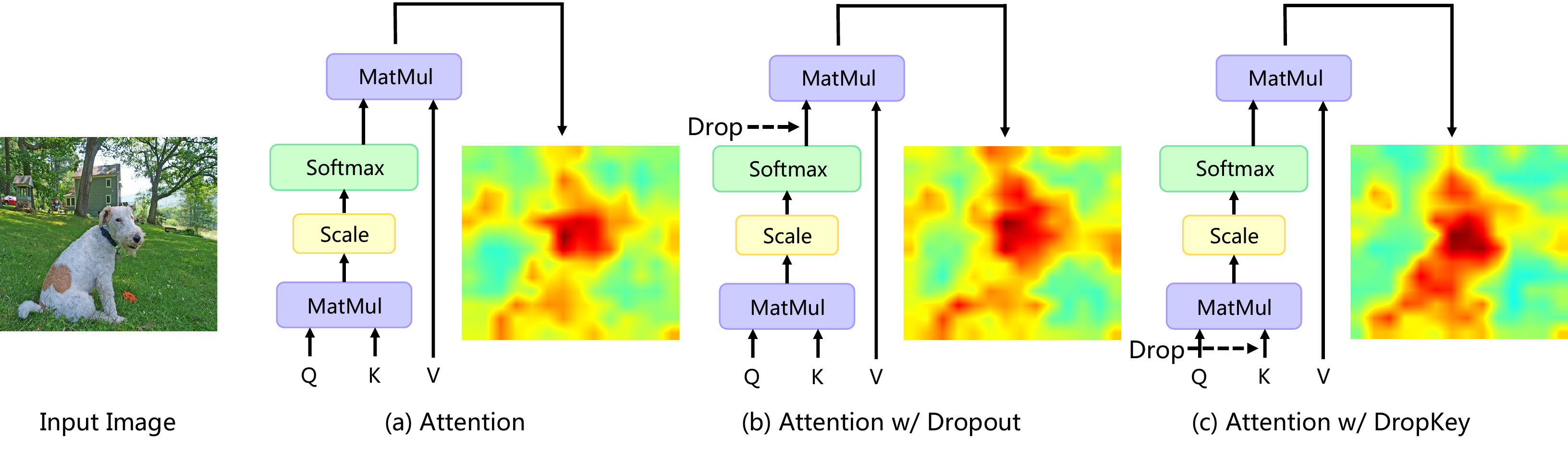}
  \caption{Comparison between the proposed DropKey and existing vanilla dropout techniques in self-attention layers for ViTs. (a) Self-attention without dropout, which suffers overfittings problem to local patch; (b) Self-attention with vanilla dropout, which regularizes the attention weights but still overfits specific patterns; (c) Self-attention with our DropKey, which overcomes prior problems and improves the model to capture vital information in a global manner. Best viewed in color.  
  }
\label{fig:motivation}
\end{figure}

\paragraph{What to drop in self-attention layer} Different from dropping attention weights as in the vanilla design, we propose to set the Key as the dropout unit, which is essential input of self-attention layer and significantly affects the output. This moves the dropout operation forward before calculating the attention matrix as shown in Fig.~\ref{fig:motivation} (c) and yields a novel dropout-before-softmax scheme. This scheme regularizes attention weights and keeps their probability distribution at the same time, which intuitively helps penalize weight peaks and lift weight foots. We theoretically verify this property via implicitly introducing an adaptive smoothing coefficient for the attention operator from the perspective of gradient optimization by formulating a Lagrange function. With the dropout-before-softmax scheme, self-attention layers can capture vital information in a global manner, thus overcoming the overfittings problem to specific patterns occurred in the vanilla dropout and enhancing the model generalizability as visualization of feature map in Fig.~\ref{fig:motivation} (c). For the training phase, this scheme can be simply implemented by swapping the operation order of softmax and dropout in vanilla design, which provides a general way to effectively enhance ViTs. For the inference phase, we conduct an additional finetune phase to align the expectations to the training phase, further improving the performance.

\paragraph{How to schedule the drop ratio} In contrast to exploiting a constant drop ratio for all layers, we present a new linear decreasing schedule that gradually decreases the drop ratio along the stack of self-attention layers. This schedule leads to a high drop ratio in shallow layers while the low one in deep layers, thus avoiding overfittings to low-level features and preserving sufficient high-level semantics. We experimentally verify the effectiveness of the proposed decreasing schedule for drop ratio to stable the training phase and improve the robustness. 

\paragraph{Whether need to perform structured drop} Inspired by the DropBlock~\citep{dropblock} method for CNNs, we implement two structured versions of the dropout operation for ViTs: the block-version dropout that drops keys corresponding to contiguous patches in images or feature maps; the cross-version dropout that drops keys corresponding to patches in horizontal and vertical stripes. We conduct thorough experiments to validate their efficacy and find that the structure trick useful for CNN is not essential for ViT, due to the powerful capability of ViT to grasp contextual information in full image range. 

Given exploration on the above three aspects, we present a novel DropKey method that utilizes Key as the drop unit and decreasing schedule for drop ratio. In particular, DropKey overcomes drawbacks of the vanilla dropout technique for ViTs, improving performance in a general and effective way. Comprehensive experiments on different ViT architectures and vision tasks demonstrate the efficacy of DropKey. Our contributions are in three folds: First, to our best knowledge, we are the first to theoretically and experimentally analyze dropout technique for self-attention layers in ViT from three core aspects: drop unit, drop schedule and structured necessity; Second, according to our analysis, we present a novel DropKey method to effectively improve the dropout technique in ViT. Third, with DropKey, we improve multiple ViT architectures to achieve new SOTAs on various vision tasks.

\section{Related Work}

\paragraph{Vision Transformers}
Inspired by the Transformer architectures in NLP, some works introduce attention block to replace convolution layers to model long-range dependencies~\citep{vit,rethinking,pyramid,cvt,end,pct}.
ViT\citep{vit} is the pioneering work which splits an image to non-overlapping patches and then these patches are fed to a transformer to evaluate attention scores. Prior works mainly focus on patch division, architecture design and tasks extension. For example, CvT\citep{cvt} proposes a hybrid architecture to mix convolutional with attention layers to introduce local inductive bias. \citep{swin,twin,mvit} proposes to reduces the computation cost.
Despite the progress, previous works ignore the dropout technique in self-attention layer, which play an important role to keey good generalizability.

\paragraph{Dropout}

Dropout is a common technique for improving the  generalizaibility of neural networks,
\emph{e.g.}, CNNs~\citep{cnn}, RNNs~\citep{rnn} and GNNs~\citep{gnn}.
For CNNs, \citep{dropblock} points out the lack of success of dropout for convolutional layers is due to dense information flowing and then proposes a form of structured Dropout. DropConnect~\citep{dropconnect} randomly masks a subset of weights within the model. For RNNs, the first prominent research on dropout is presented in ~\citep{rnnv}, which is equipped with a learnable dropout rate.

For GNNs, to alleviate over-fitting and over-smoothing issues, DropEdge~\citep{dropedge} randomly removes part of edges in the input stage at each training epoch and theoretically demonstrates the effect of the proposed method. For ViTs, most original publicly code of multiple popular vision transformer simply intuitively apply dropout operation on attention matrix.

\paragraph{DropAttention}
\citep{dropattention} presents a novel way for transformer in NLPs.
It refers to perform randomly drop on the attention matrix.
There are two two main forms of applying DropAttention for the transformer training, DropAttention-unit and DropAttention-elements.
Similar to the standard Dropout~\citep{dropout}, DropAttention-unit randomly drops the unit, that is, these dropped token will not be used as attend patches for each query patch.
As the general form of DropAttention-unit, DropAttention-elements randomly drops elements in attention weights matrix which is similar to DropConnect~\citep{dropconnect}.
Different from standard Dropout~\citep{dropout}, DropAttention introduce re-normalize to guarantee the sum of attention weights remain 1 and help training process to be more steady.
However, there is no further discussion and theoretical analysis on this phenomenon.

\section{Method}
Our generic DropKey is inspired by DropAttention and proposed for self-attention operator in vision transformer.
The main idea of DropKey is to adaptively adjust attention weight to obtain a smoother attention vector.
In this section, we start by introducing the theoretical explanation of DropKey and then the implementation will be explained in detail.

\subsection{Methodology}
\label{sec:methodology}
As discussed in the above section, transformer-based model tends to rely on local features rather than general global information.
To alleviate this issue, we propose to reduce local-bias by encouraging models to learn a smoother attention weight for each patch.
To this end, we attempt to reduce the attention weight of the patch which has a large attention weight, and vice versa.
Nevertheless, it is tedious and difficult to achieve this by explicitly setting rules.
In this work, we found DropKey achieves the above implicitly by normalizing the attention vector that has performed the dropout operation.
Specifically, by given an image $I \in \mathbb{R}^{H \times W \times C}$, where $H$ denotes height, $W$ denotes width and $C$ denotes channels, Vision Transformer architecture starts by dicing it into $n_{h} \times n_{w}$ patches $x\in\mathbb{R}^{ n_{c}}$.
Then, $x$ is used as input of self- attention layer and the output $o$ can be computed as follows
\begin{equation}
\label{eq:equ}
o = \sum_{j=1}^{n_{h}n_{w}}(\frac{d_{j}p_{j}}{\sum_{j=1}^{n_{h}n_{w}}d_{j}p_{j}})v_{j}
\end{equation}
\begin{equation}
\label{eq:softmax}
p_{j} = \frac{exp(\frac{qk_{j}^{T}}{scale})}{\sum_{j=1}^{n_{h}n_{w}}exp(\frac{qk_{j}^{T}}{scale})}
\end{equation}
where $q{=}\mathcal{F}_{q}(x), k{=}\mathcal{F}_{k}(x), v{=}\mathcal{F}_{v}(x)$, and $q_{i}$, $k_{j}$, $v_{j}$ denote query of $i^{th}$ patch, key of $j^{th}$ patch, value of $j^{th}$ patch and $o$ denotes the output of one patch.
$\mathcal{F}_{q}(\cdot)$, $\mathcal{F}_{k}(\cdot)$, $\mathcal{F}_{v}(\cdot)$ denote projection layers with weights of dimensions $n_{c} \times n_{c}$.
$scale$ denotes scaling factor and is set to $\sqrt{n_{c}}$.
$d$ denotes drop ratio and $d_{j} \sim Bernoulli(1-d)$.
Here, for convenience, we focus on one head and omit the index of patch.
By performing re-normalize, DropKey has the ability to adaptively adjust attention weight to smoother.
Next, we provide theoretical analysis of our method to demonstrate its effectiveness.

Here, we formulate the expectation output of model via introducing DropKey in training stage as:
\begin{equation}
\label{eq:E}
\begin{aligned}	 
		\mathbb{E}_{d_{j},1 \leq j \leq n_{h}n_{w}}[o] &= \mathbb{E}_{d_{j},1 \leq j \leq n_{h}n_{w}}[\sum_{j=1}^{n_{h}n_{w}}(\frac{d_{j}p_{j}}{\sum_{j=1}^{n_{h}n_{w}}d_{j}p_{j}})v_{j}] \\
		& = \sum_{j=1}^{n_{h}n_{w}}\mathbb{E}_{d_{j},1 \leq j \leq n_{h}n_{w}}[\frac{d_{j}}{\sum_{j=1}^{n_{h}n_{w}}d_{j}p_{j}}]p_{j}v_{j} = \sum_{j=1}^{n_{h}n_{w}}c_{j}p_{j}v_{j}
	 \end{aligned}
\end{equation}
where $c_{j} = \mathbb{E}_{d_{j},1 \leq j \leq n_{h}n_{w}}[\frac{d_{j}}{\sum_{j=1}^{n_{h}n_{w}}d_{j}p_{j}}]>0$ is a smoothing coefficient which is related to $d_{j}$ and $p_{j}$.

Note that $c$ in Equ~\ref{eq:E} can be considered to add an additional smoothing prior to distribution $p$ , i.e., $c_{s} < c_{t}$ when $p_{s} > p_{t}$, $1\le s,t\le n_{h}n_{w}$.
The specific proof is as follows:

\begin{equation}
\label{eq:c}
\begin{aligned}	 
		c_{s} - c_{t} &=\mathbb{E}_{d_{j},1 \leq j \leq n_{h}n_{w}}[\frac{d_{s}}{\sum_{j=1}^{n_{h}n_{w}}d_{j}p_{j}}]-\mathbb{E}_{d_{j},1 \leq j \leq n_{h}n_{w}}[\frac{d_{t}}{\sum_{j=1}^{n_{h}n_{w}}d_{j}p_{j}}]\\
		& = d(1-d)\mathbb{E}_{d_{j},1 \leq j \leq n_{h}n_{w}}[\frac{-1}{\sum_{j\ne s,t}^{n_{h}n_{w}}d_{j}p_{j}+p_{t}}]+d(1-d)\mathbb{E}_{d_{j},1 \leq j \leq n_{h}n_{w}}[\frac{1}{\sum_{j\ne s,t}^{n_{h}n_{w}}d_{j}p_{j}+p_{s}}]\\
		& = d(1-d)\mathbb{E}_{d_{j},1 \leq j \leq n_{h}n_{w}}[\frac{1}{\sum_{j\ne s,t}^{n_{h}n_{w}}d_{j}p_{j}+p_{s}}-\frac{1}{\sum_{j\ne s,t}^{n_{h}n_{w}}d_{j}p_{j}+p_{t}}]<0
	 \end{aligned}
\end{equation}
It is natural to find $c$ serves as a factor to implicitly encourage the model properly to reduce the consideration of the patch with large $p$ and improve the effectiveness of the patch with small attention weight. 
Meanwhile, $c$ can be adjusted adaptively according to the distribution of samples in the training stage without any manual design.
Nevertheless, a noteworthy problem is that since we removed DropKey in the inference phase, the output expectations in training and inference stage are inconsistent, which will decrease the performance.
In Section~\ref{sec:align}, we propose two methods to alleviate this problem.

Having explored the implicit regularization effects of DropKey, we also demonstrate it from the perspective of gradient optimization.
For simplicity, we start by considering a simple but universal optimization objective as follows:
\begin{equation}
\label{eq:simple}
\underset{p_{j},v_{j}}{min}\ \frac{1}{2}||\sum_{j=1}^{N}p_{j}v_{j}-y||^{2}\qquad
\mathrm{s.t.}\quad p_{j}>0,\sum_{j=1}p_{j}=1,1\le j\le N
\end{equation}
where $p_{j}\in \mathbb{R}$, $v_{j}\in \mathbb{R}^{r}$ denotes learnable parameters and $y\in \mathbb{R}^{r}$ denotes target.
Here, $p_{j}$ and $v_{j}$ can be considered as attention weight and value in attention mechanism.
Naturally, $v_{j}$ can be decomposed into two directions, which are the same direction as $y$ and the direction perpendicular to $y$:
\begin{equation}
\label{eq:decompose}
v_{j} = \beta_{j}e+\alpha_{j}
\end{equation}
where $\beta$ is a scalar, $e\in \mathbb{R}^{r}$ is the unit vector of $y$ and $\alpha_{j}$ is the component perpendicular to $y$.
Similarly, we can rewrite $y$ to $Me$ where $M=||y||$.
Then, the Lagrange function of Equ~\ref{eq:simple} can be formulated as follows:
\begin{equation}
\label{eq:Lagrange}
L(p_{j},v_{j},\lambda)=\frac{1}{2}||\sum_{j=1}^{N}p_{j}v_{j}-y||^{2}-\lambda(\sum_{j=1}^{N}p_{j}-1)
\end{equation}

In order to analysis the gradient, we take the partial derivatives of $p_{j}$ and $v_{j}$:
\begin{equation}
\label{eq:partial_p}
\frac{\partial L}{\partial p_{j}} = (v - y)^{T}v_{j}-\lambda = (\sum_{j=1}^{N}p_{j}\beta_{j}-M)\beta_{j}+(\sum_{j=1}^{N}p_{j}\alpha_{j})^{T}\alpha_{j}-\lambda
\end{equation}
\begin{equation}
\label{eq:partial_v}
\frac{\partial L}{\partial v_{j}} = p_{j}(v-y) = p_{j}(\sum_{j=1}^{N}p_{j}\beta_{j}-M)e+p_{j}(\sum_{j=1}^{N}p_{j}\alpha_{j})
\end{equation}
Here, for simplicity, we note $v=\sum_{j=1}^{N}p_{j}v_{j}=(\sum_{j=1}^{N}p_{j}\beta_{j})e+\sum_{j=1}^{N}p_{j}\alpha_{j}$.
From the chain rule of backpropagation, we also have:
\begin{equation}
\label{eq:partial_betaalpha}
\frac{\partial L}{\partial \beta_{j}} = p_{j}(\sum_{j=1}^{N}p_{j}\beta_{j}-M)
\qquad
\frac{\partial L}{\partial \alpha_{j}} = p_{j}(\sum_{j=1}^{N}p_{j}\alpha_{j})
\end{equation}
Equ~\ref{eq:partial_betaalpha} indicates that the gradient $\frac{\partial L}{\partial \beta_{s}} < \frac{\partial L}{\partial \beta_{t}} $ when $p_{s}>p_{t}$ and this backward propagation properties would lead $\frac{\partial L}{\partial p_{s}} < \frac{\partial L}{\partial p_{t}}$.
By subtracting $\frac{\partial L}{\partial p_{s}}$ and $\frac{\partial L}{\partial p_{t}}$, we will have:
\begin{equation}
\label{eq:partial_sub}
\frac{\partial L}{\partial p_{s}} - \frac{\partial L}{\partial p_{t}}=(\sum_{j=1}^{N}p_{j}\beta_{j}-M)(\beta_{s}-\beta_{t})+(\sum_{j=1}^{N}p_{j}\alpha_{j})^{T}(\alpha_{s}-\alpha_{t}) 
\end{equation}
Since all parameters are randomly initialized, it can be assumed that the initial solution is far from the optimal solution, that is, $p^{(0)}_1=p^{(0)}_2= \cdots =p^{(0)}_N=\frac{1}{N}$, $\vert \beta^{(0)}_j \vert \ll M, 1 \leq j \leq N$.
Consequently, the first additive term plays the leading role in Equ~\ref{eq:partial_sub}.
Due to $\frac{\partial L}{\partial \beta_{s}} < \frac{\partial L}{\partial \beta_{t}}$, the update speed of $\beta_{s}$ is faster than $\beta_{t}$ and then lead $\beta_{s}>\beta_{t}$, $\frac{\partial L}{\partial p_{s}} < \frac{\partial L}{\partial p_{t}}$.
Based on the above analysis, it can be concluded that a larger $p$ would promote $\beta$ larger, while a larger $\beta$ would further promote $p$ larger.
Finally, the output of model would be controlled by a few sparse blocks.
Conversely, DropKey avoids suffering local-bias by introducing the parameter $c$ to enforce distribution $p$ to be smoother.

\subsection{Implementation}

\paragraph{What to drop?}Different from DropAttention, we integrate Dropout and re-normalize into one stage by dropping key rather than weight.
At each training iteration, DropKey masks a certain rate of keys of the input key map by random.
It is worth noting that we generate masked key map for each query, instead of sharing the same  masked key map for all query vectors.
Specifically, given the query $Q \in \mathbb{R}^{n_{h}n_{w}\times n_{c}}$, key $K \in \mathbb{R}^{n_{h}n_{w}\times n_{c}}$ and value $V \in \mathbb{R}^{n_{h}n_{w}\times n_{c}}$ of a feature map, it first computes the dot products of the query with all keys and divide each by scaling factor $scale=\sqrt{n_{c}}$.
Subsequetly, we randomly generate a mask matrix $D\in \mathbb{R}^{1\times(n_{h}n_{w}\times n_{h}n_{w})}$ with drop ratio $d$ to enforce some elements of the similarity matrix to be -inf.
The formulation of $D$ as follows:
\begin{equation}
\label{eq:mask}
d_{j}=\left\{
\begin{array}{rcl}
0 & & with\ probability\ 1-d \\
-\infty & & with\ probability\ d\\
\end{array} \right.
\end{equation}

\begin{figure}
	\centering
	\begin{minipage}{0.55\linewidth}
		\begin{algorithm}[H]
        \caption{{Attention with DropKey code}}
        \label{alg:att_da_dp}
        \definecolor{codeblue}{rgb}{0.25,0.5,0.25}
        \lstset{
        	backgroundcolor=\color{white},
        	basicstyle=\fontsize{6.5pt}{6.5pt}\ttfamily\selectfont,
        	numbers=left, 
        	columns=fullflexible,
        	breaklines=true,
        	captionpos=b,
        	commentstyle=\fontsize{7.2pt}{7.2pt}\color{codeblue},
        	keywordstyle=\fontsize{7.2pt}{7.2pt},
        	emph={m_r = torch.ones_like(attn) * mask_ratio, attn = attn + torch.bernoulli(m_r) * -1e12},
        	emphstyle=\color{red}
        }
        \begin{lstlisting}[language=python]
        # N: token number, D: token dim
        # Q: query (N, D), K: key (N, D), V: value (N, D)
        # use_DropKey: whether use DropKey
        # mask_ratio: ratio to mask 
        
        def Attention(Q, K, V, use_DropKey, mask_ratio)
            attn = (Q * (Q.shape[1] ** -0.5)) @ K.transpose(-2, -1)
            
            # use DropKey as regularizer
            if use_DropKey == True:
                m_r = torch.ones_like(attn) * mask_ratio
                attn = attn + torch.bernoulli(m_r) * -1e12
                
            attn = attn.softmax(dim=-1)
            x = attn @ V
            return x
        \end{lstlisting}
        \end{algorithm}
	\end{minipage}
	\begin{minipage}{0.44\linewidth}
		\centering
		\includegraphics[width=0.9\linewidth]{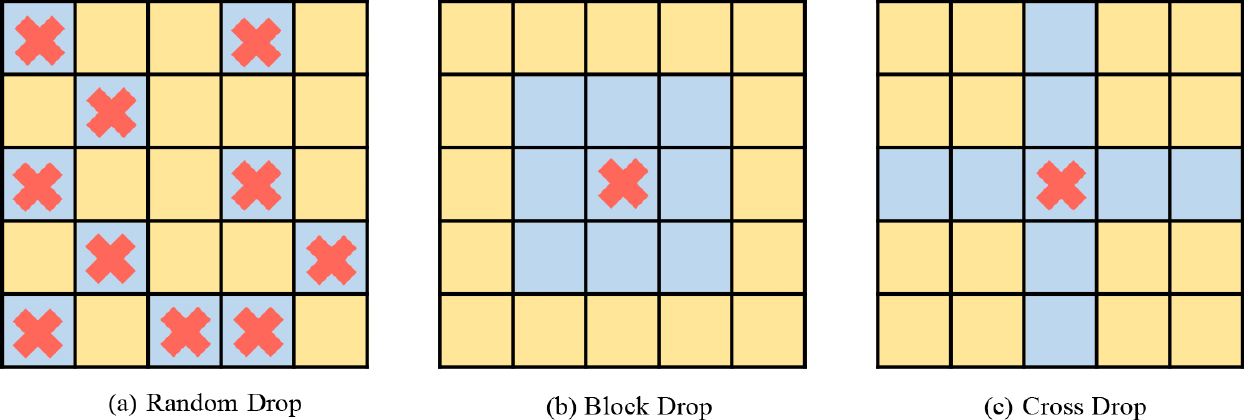}
		\caption{ Mask sampling in DropKey, DropKey-Block and DropKey-Cross. The yellow patches are used as attend key to interact with a query and blue patches are dropped. Red symbol denotes the valid seed and the window size of DropKey-Block and DropKey-Cross is 3 and 1, respectively.}
		\label{fig:dropStyle}
	\end{minipage}

\end{figure}

Finally, the attention weight matrix is computed by given masked similarity matrix as input.
Formally, we compute the outputs of a patch as:
\begin{equation}
\label{eq:attn_2}
o =\sum_{j=1}^{n_{h}n_{w}}p_{j}v_{j} 
\end{equation}
\begin{equation}
\label{eq:droppatch_1}
p_{j} = \frac{exp(d_{j}+ \frac{qk_{j}^{T}}{scale})}{\sum_{j=1}^{n_{h}n_{w}}exp(d_{j}+\frac{qk_{j}^{T}}{scale})}
\end{equation}
where $q_{i}$, $k_{j}$, $v_{j}$ denote query of $i^{th}$ patch, key of $j^{th}$ patch, value of $j^{th}$ patch.
It is not difficult to find that Equ~\ref{eq:attn_2} and Equ~\ref{eq:droppatch_1} is the same equivalent form of Equ~\ref{eq:equ} and Equ~\ref{eq:softmax}.
In particular, as shown in Algorithm~\ref{alg:att_da_dp}, we only need to add two lines of code to pure attention to implement DropKey. Obviously, compared with Dropout, DropKey only changes the position of introducing mask.

\paragraph{Scheduled Drop Ratio}
Vision transformer always consists of few self-attention blocks to gradually learn high-dimensional features.
Generally, early layers operates low-level visual features and deeper layers aim to model spatially coarse but complex information.
Hence, we attempt to set smaller drop ratio for deeper layers to avoid missing important objects-relevant information.
Specifically, instead of stochastically dropping with some fixed probability in Dropout, we gradually decrease number of dropped keys over the layers during training stage.
We find that this scheduled drop ratio not only work well for DropKey, but also significantly improves the performance of Dropout.

\paragraph{Structured DropKey}
Although structured drop has been detailedly explored in convolution network, there is no work to study the impact of structured drop on vision transformer.
In this section, we implement two structured forms of DropKey which are named DropKey-Block and DropKey-Cross (see Figure~\ref{fig:dropStyle}).
For DropKey-Block, inspired by DropBlock~\citep{dropblock}, we drop contiguous patches in a square-shape window of feature maps.
Actually, we re-set drop ratio $d_{block}$ for each patch as:
$d_{block} = \frac{d}{s^{2}}\frac{n_{h}n_{w}}{(n_{h}-s+1)(n_{w}-s+1)}$
where $d$ denotes the probability of dropping a patch in DropKey.
The valid seed region is $(n_{h}-s+1)(n_{w}-s+1)$ where $s$ denotes the size of window and $n_h,n_w$ denotes height,weight of feature map.

The cross-shape window attention is proved to achieve strong modeling capability~\citep{cross} which also indicates the information in this structure is correlated.
For DropKey-Cross, we discard features in a cross window to prevent information flow through self-attention.
In our implementation, we drop the rows and columns of valid seed and re-set the drop ratio $d_{cross}$ for each patch as:
$d_{cross} = \frac{d}{s(n_{h}+n_{w}-s)}\frac{n_{h}n_{w}}{(n_{h}-s+1)(n_{w}-s+1)}$. Note that the formulations for block- and cross-version are only the approximation, because there will be some overlapped when perform drop.

\paragraph{Align Expectation}
\label{sec:align}
As mentioned in Section~\ref{sec:methodology}, the misaligned expectations have a certain negative impact on the model, so we attempt to use two methods to align the expectation.
The first one is to use Monte Carlo method to estimate $c$.
We perform multiple random drop and calculate the attention weight matrix after each drop operation.
Finally, the average of calculated multiple weight matrices is applied as the input for the next step. 
For the second one, we take the inspiration from ~\citep{bias} and propose to finetune the model without DropKey, as an extra stage after DropKey training. We experimentally verify that the second strategy performs better.

\section{Experiments}
We conduct experiments on three tasks,
image classification, object detection, human-object interaction detection and human mesh recovery, to show the efficacy and generalizability of our DropKey for improving ViTs.
\subsection{Image Classification}

\paragraph{Datasets}
We conduct the following experiments with T2T~\citep{t2t} and VOLO~\citep{volo} for image classification on CIFAR10~\citep{cifar10}, CIFAR100~\citep{cifar100} and ImageNet~\citep{imagenet}.
a) We conduct the ablation study to demonstrate the effects of introducing modified DropAttention to three T2T and three VOLO backbone architectures on CIFAR10 and CIFAR100. 
b) We validate the modified DropAttention by training T2T and VOLO from scratch on ImageNet.

\paragraph{Training}
By default, models are trained with random initialization on all datasets.
Due to the missing training setting for T2T~\citep{t2t} and VOLO~\citep{volo}, we conduct grid search on learning rate and epoch number to make the vanilla backbones achieve the best performance.
All implementation specifics can be seen in Appendix.
For ImageNet, our training recipe follows \citep{t2t,volo}.
Specifically, we set 50 epoches and keep learning rate as $10^{-5}$ for finetune stage for all datasets.

\subsubsection{Ablations on CIFAR10 and CIFAR100}
\begin{figure}
  \centering
  \includegraphics[width=\textwidth]{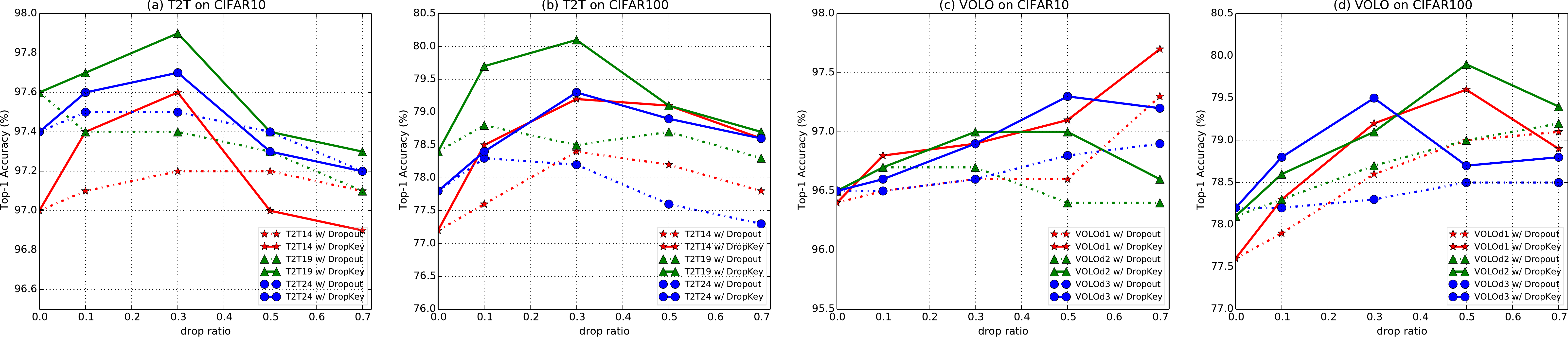}

  \caption{Comparison of models with Dropout or DropKey on CIFAR10 and CIFAR100.}
 \label{fig:dropkey}

\end{figure}

\paragraph{Does DropKey boost generalization?}
T2T~\citep{t2t} and VOLO~\citep{volo} are the widely used vision transformer architecture for image recognition.
However, the DropAttention applied in these architectures does not apply re-normalize operation.
In the following experiments, we plug in the DropKey to the family of T2T/VOLO and compare results with the version that use Dropout.
The results in Figure~\ref{fig:dropkey} indicate DropKey can consistently increase the performance of various architectures on CIFAR10 and CIFAR100.
On CIFAR100 with T2T19, for example, DropKey gains 1.7\% improvement for the adaptive coefficient yields further improvement.
Additionally, we found that drop ratio is also an essential factor affecting performance.
Take T2T14 with DropKey as an example, the accuracy initially increases with increasing drop ratio until it achieves its peak accuracy of 79.2\% at drop ratio = 0.3 and it declines upon further increase of the drop ratio.
These suggest that the introduction of DropKey would be beneficial to the network for avoiding suffering overfits. 
However, a larger drop ratio can lead to an inability to capture valid information about the object. 
Similar to DropKey, the performance of Dropout is also related to drop ratio.

\paragraph{How does the choice of scheduled strategy of drop ratio impact accuracy?}
The ablation in Table~\ref{tab:sheduled} analyzes
the accuracy of DropKey by adjusting the strategy of setting drop ratio.
Specifically, we validate three strategy as follows: a) Constant: drop ratio is constant over self attention layers. b) Scheduled$\uparrow$: drop ratio is linearly increased over self attention layers. c) Scheduled$\downarrow$: drop ratio is linearly decreased over self attention layers.
Firstly, we find that Scheduled$\downarrow$ outperform Constant whether using re-normalize or not.
Secondly, introducing drop ratio with Scheduled$\uparrow$ would seriously affect the performance of the model and even perform worse than the pure vision transformer.
This phenomenon is in line with our expectation, since deeper layers often contain high-level semantic information which is essential to perform classification.
A large drop ratio in deep layer will increase the risk of losing important features which would make model difficult to converge to a generalized solution.
Hence, we set drop ratio with Scheduled$\downarrow$ in all following experiments.
\begin{table}
\scriptsize
  \caption{ Comparison of models with Dropout or DropKey on CIFAR10 and CIFAR100 when introducing different scheduled strategy of drop ratio.
  The drop ratio is set as 0.3 for all models.}

  \label{tab:sheduled}
  \centering
  \begin{tabular}{ccccccc}
    \toprule
     \multirow{2}{*}{Model} & \multicolumn{3}{c}{CIFAR10} & \multicolumn{3}{c}{CIFAR100}\\
    \cmidrule(r){2-4} \cmidrule(r){5-7}
    & Scheduled$\uparrow$ & Constant & Scheduled$\downarrow$ & Scheduled$\uparrow$ & Constant & Scheduled$\downarrow$\\
    \cmidrule(r){1-7}
    \cmidrule(r){1-7}
    T2T14~\citep{t2t} + Dropout & 96.7 & 96.9 & \textbf{97.2} &77.0 &77.8 &\textbf{78.4}\\
    T2T14~\citep{t2t} + DropKey &97.0 & 97.4& \textbf{97.6}&77.2 &78.6 &\textbf{79.2}\\
    T2T19~\citep{t2t} + Dropout &96.6 & 97.1 & \textbf{97.4}&77.3 &78.1 &\textbf{78.5}\\
    T2T19~\citep{t2t} + DropKey &97.1 & 97.2& \textbf{97.9}&78.0 &78.4 &\textbf{80.1}\\
    T2T24~\citep{t2t} + Dropout &96.4 &97.0 & \textbf{97.5}&76.7 & 77.3 &\textbf{78.2}\\
    T2T24~\citep{t2t} + DropKey &96.6 & 97.2 & \textbf{97.7}& 76.9&78.3 &\textbf{79.3}\\
    \cmidrule(r){1-7}
    \cmidrule(r){1-7}
    VOLOd1~\citep{volo} + Dropout &95.8 & 96.4& \textbf{96.6}& 77.3& 78.2&\textbf{79.0}\\
    VOLOd1~\citep{volo} + DropKey &96.2 & 96.7& \textbf{97.1}&78.1 & 78.6&\textbf{79.6}\\
    VOLOd2~\citep{volo} + Dropout & 95.6 & 96.1& \textbf{96.4}& 77.0& 78.4&\textbf{79.0}\\
    VOLOd2~\citep{volo} + DropKey & 95.7& 96.6 & \textbf{97.0}&77.7 & 79.0&\textbf{79.9}\\
    VOLOd3~\citep{volo} + Dropout &96.0 &96.3 & \textbf{96.6}&77.1 & 77.9&\textbf{78.5}\\
    VOLOd3~\citep{volo} + DropKey & 96.2& 96.4 & \textbf{96.9} &77.4 & 78.1&\textbf{78.7}\\
    \bottomrule
  \end{tabular}
\end{table}

\begin{table}
  \caption{ Comparison of different structured drop on CIFAR100. $s$ denotes window size and drop ratio is set as 0.3 for all models.}
  \label{tab:structure_drop}
  \centering
  \scriptsize
  \begin{tabular}{cccccc}
    \toprule
     \multirow{2}{*}{Model} & \multirow{2}{*}{Random}& \multicolumn{2}{c}{Block} & \multicolumn{2}{c}{Cross}\\
    \cmidrule(r){3-4}
    \cmidrule(r){5-6} 
     & & $s$=3 & $s$=5& $s$=1 & $s$=3 \\
    \cmidrule(r){1-6}
    \cmidrule(r){1-6}
    T2T14~\citep{t2t} + Dropout & \textbf{78.4}& 78.1 & 77.1& 77.9 & 76.8\\
    T2T14~\citep{t2t} + DropKey &\textbf{79.2}& 78.7 & 77.5 & 78.6 & 77.4\\
    VOLOd1~\citep{volo} + Dropout & \textbf{78.6} & 78.0 & 76.7 & 77.6 & 76.4 \\
    VOLOd1~\citep{volo} + DropKey & \textbf{79.2} & 78.5 & 77.5 & 78.3 & 77.1\\
    \bottomrule
  \end{tabular}
\end{table}

\paragraph{Is Dropkey helpful to capture global information?}
The contribution of DropKey is to generate a smoother attention weight matrix to encourage models to focus on global features. 
Class token has been proved to be effective in aggregating the information of the whole image, so we measure the smoothness of the attention distribution by calculating the entropy of its attention weight vector.
Specifically, we define the entropy of the attention vector of class token as follows
$E_{i} =-\sum_{j}p_{j}^{i}\log p_{j}^{i}$
where $p_{j}^{i}$ denotes the attention weight of $j^{th}$ attending patch in $i^{th}$ head.

\begin{figure}[t!]
	\centering
	\begin{minipage}{0.44\linewidth}
		\centering
		\includegraphics[width=0.8\linewidth]{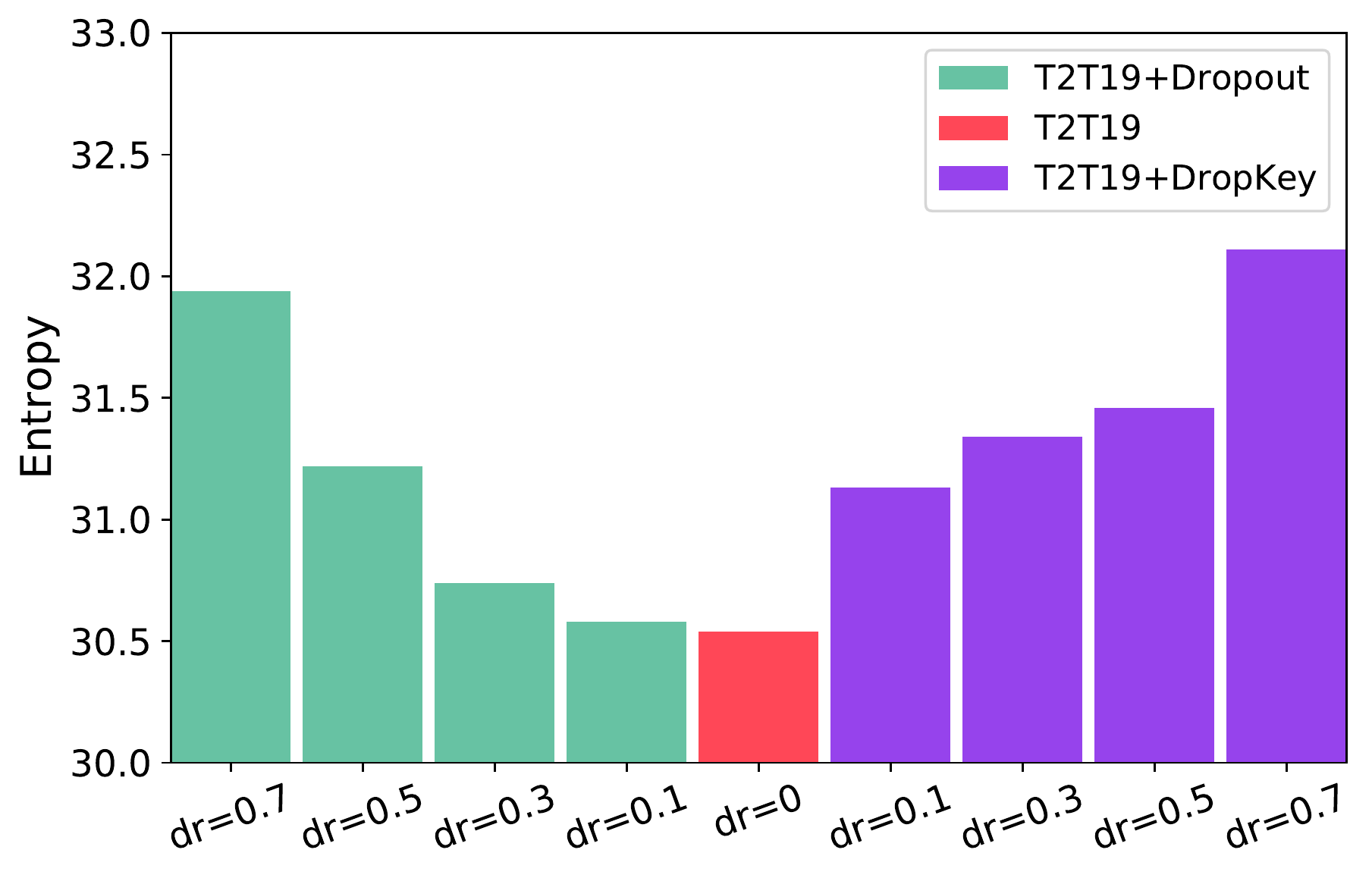}

		\caption{ The histogram of entropy on CIFAR100.
		dr denotes drop ratio.
		}
		\label{fig:entropy}
	\end{minipage}
	\begin{minipage}{0.54\linewidth}
		\centering
		\includegraphics[width=0.9\linewidth]{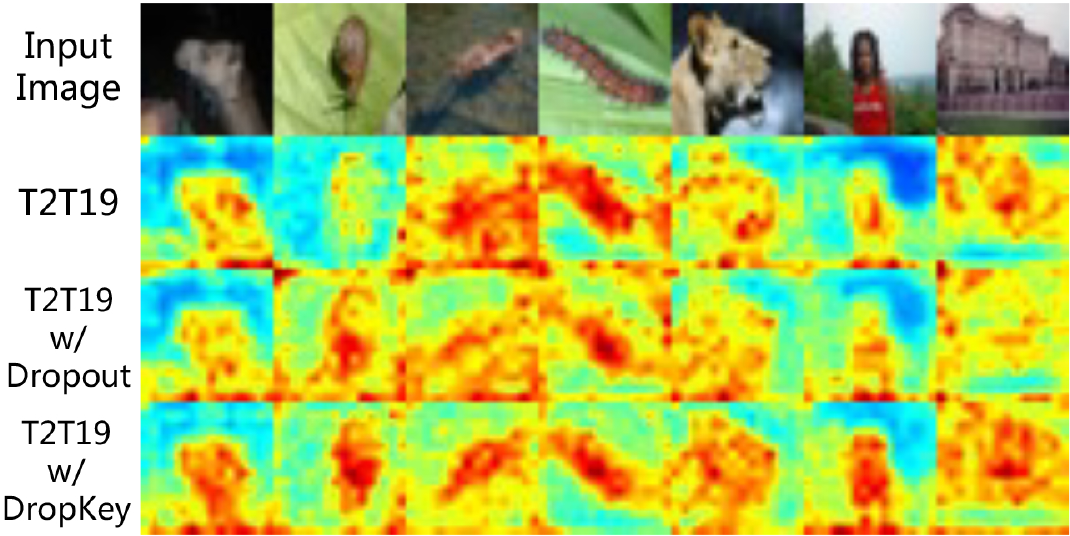}

		\caption{Visualization on CIFAR100.}
		\label{fig:vis_cifar100}
	\end{minipage}
\end{figure}

Note that a small entropy denotes that this head focuses on sparse patches.
For convenience, the mean of entropy of multi-head as $E$ is reported in Figure~\ref{fig:entropy}.
Clearly, it reads that while both Dropout and DropKey are able to improve the value of entropy, the improvement by DropKey is more significant.
Meanwhile, when the drop rate increases, the entropy increase accordingly.
In addition, to further verify the necessity of the adaptive coefficient for capturing general global information,  we visualize the attention map (Figure~\ref{fig:vis_cifar100}) from class token in the last transformer layer.
Obviously, the model with DropKey smoothly assigns attention weight to the area related to the instance.

\paragraph{Is Dropkey help model robust to occlusions?}
\label{sec:occ}
To further verify that DropKey can alleviate local-bias problems, we study whether vision transformer with DropKey perform robustly in occluded scenarios, where some content of the image is missing.
Specifically, a subset of patches is randomly selected and dropped before input to self-attention layer.
In this experiment, with the drop ratio found in the manuscript, we randomly drop 10\%, 30\%, 50\%, 70\% and 90\% of the patches to test the performance of trained models.
For convenience, we define information loss as the ratio of dropped and total patches.
To eliminate the influence of random occlusions, we report the mean of accuracy across 5 runs.
The results reported in Table~\ref{tab:occlusions} show significantly robust performance of the model with DropKey against its with Dropout trick.
For example, T2T24+Dropout achieves 75.7\% accuracy in comparison to T2T24+DropKey which obtains 78.0\% accuracy when 30\% of the patches are removed.
A surprising phenomenon can be observed that when 90\% of the image information is randomly dropped, T2T19+DropKey and T2T24+DropKey still exhibits 32.7\% and 29.7\% accuracy, respectively.
Consequently, compared with Dropout, models with DropKey show significant robustness to the content removal.
\begin{table}
  \caption{ Robustness against occlusion in images is studied under T2T19 and T2T24 on CIFAR100.
  The drop ratio is set as 0.3 for all models.}
  \label{tab:occlusions}
  \centering
  \begin{tabular}{c|cccccc}
    \toprule
     Information Loss & 0.0 & 0.1 & 0.3 & 0.5 & 0.7 & 0.9\\
    \cmidrule(r){1-7}
     T2T19~\citep{t2t} & $78.4$ & $77.6^{\textcolor{red}{\textbf{-0.8}}}$ & $75.6^{\textcolor{red}{\textbf{-2.8}}}$ & $73.1^{\textcolor{red}{\textbf{-5.3}}}$ & $65.1^{\textcolor{red}{\textbf{-13.3}}}$ & $28.2^{\textcolor{red}{\textbf{-50.2}}}$\\
     T2T19~\citep{t2t} + Dropout & $78.5$ & $77.4^{\textcolor{red}{\textbf{-1.1}}}$ & $76.5^{\textcolor{red}{\textbf{-2.0}}}$ & $73.9^{\textcolor{red}{\textbf{-4.6}}}$ & $66.0^{\textcolor{red}{\textbf{-12.5}}}$ & $30.2^{\textcolor{red}{\textbf{-48.3}}}$\\
     T2T19~\citep{t2t} + DropKey & $80.1$ & $79.4^{\textcolor{red}{\textbf{0.7}}}$ & $78.2^{\textcolor{red}{\textbf{-1.9}}}$ & $75.8^{\textcolor{red}{\textbf{-4.3}}}$ & $68.7^{\textcolor{red}{\textbf{-11.4}}}$ & $32.7^{\textcolor{red}{\textbf{-47.4}}}$\\
    \cmidrule{1-7}
     T2T24~\citep{t2t} & $77.8$ & $77.1^{\textcolor{red}{\textbf{-0.7}}}$ & $75.2^{\textcolor{red}{\textbf{-2.6}}}$ & $72.7^{\textcolor{red}{\textbf{-5.1}}}$ & $63.5^{\textcolor{red}{\textbf{-14.3}}}$& $23.7^{\textcolor{red}{\textbf{-54.1}}}$\\
     T2T24~\citep{t2t} + Dropout & $78.2$ & $77.3^{\textcolor{red}{\textbf{-0.9}}}$ & $75.7^{\textcolor{red}{\textbf{-2.5}}}$ & $72.8^{\textcolor{red}{\textbf{-5.4}}}$ & $65.3^{\textcolor{red}{\textbf{-12.9}}}$ & $27.9^{\textcolor{red}{\textbf{-50.3}}}$\\
     T2T24~\citep{t2t} + DropKey & $79.3$ & $78.6^{\textcolor{red}{\textbf{-0.7}}}$ & $78.0^{\textcolor{red}{\textbf{-1.3}}}$ & $74.3^{\textcolor{red}{\textbf{-5.0}}}$ & $66.9^{\textcolor{red}{\textbf{-12.4}}}$ & $29.7^{\textcolor{red}{\textbf{-49.6}}}$\\
    \bottomrule
  \end{tabular}
\end{table}
\paragraph{Is structured drop is useful to vision transformer?}
In this section, we analyze whether structured masking is useful for vision transformer via DropKey-Block and DropKey-Cross.
Specifically, we swept over window size from 3 to 5 for DropKey-Block and from 1 to 3 for DropKey-Cross.
Compared with the DropKey, the structured drop degrade the performance of the model in all case and classification accuracy will decrease with the further increase of the window size.
One possible reason for this phenomenon is that a larger drop ratio in shallow transformer layer results in losing vital information of object.
However, a larger drop ratio is the key to avoid overfit to low-level feature.
Therefore, we encourage set a larger drop ratio rather than the introduction of structured drop.
\paragraph{How does align expectations impact accuracy?}
\label{sec:align}
As discussed above, the mismatched expectations problem arises when directly use trained model with DropKey to test.
Herein, to address above issue, we explore the impact of two methods, Monte Carlo and finetune, on the performance of network.
The results in Table~\ref{tab:align} indicates that accuracy can be improved regardless of the expectation alignment operation.
Additionally, the accuracy of Monte Carlo increases with increasing sampling iterations.
However, huge sampling iterations will lead to unacceptable computational cost.
To fairly compare with Dropout, we also finetune it with the same training hyper-parameters as DropKey and then note that finetune only brings insignificant performance improvement or even over-fitting.
In conclusion, alignment expectation of training and inference stage indeed can further improve the accuracy of the model.
\begin{table}
  \small
  \caption{ Comparison of models w/ or w/o Align Expectations on CIFAR100.
  Pure denotes the model without align expectation and drop ratio is set as 0.3 for all models.}
  \label{tab:align}
  \centering
  \begin{tabular}{ccccc}
    \toprule
     \multirow{2}{*}{Model} & \multirow{2}{*}{Pure}& \multicolumn{2}{c}{w/\ Monte Carlo} & \multirow{2}{*}{w/\ Finetune}\\
    \cmidrule(r){3-4}
     & & 2000 & 6000 &\\
    \cmidrule(r){1-5}
    \cmidrule(r){1-5}
    T2T14~\citep{t2t} + Dropout & \textbf{78.4}& - & -& \textbf{78.4}\\
    T2T14~\citep{t2t} + DropKey & 78.8& 78.9 & 79.1 & \textbf{79.2}\\
    T2T19~\citep{t2t} + Dropout & 78.5 & - & - & \textbf{78.6} \\
    T2T19~\citep{t2t} + DropKey & 79.3 & 79.5 & \textbf{80.2} & 80.1\\
    \bottomrule
  \end{tabular}
\end{table}
\subsubsection{DropKey vs Dropout on ImageNet}
We provide the accuracy under other different drop ratio for each backbone with DropKey on ImageNet, and contrast them with existing State of the Arts (SOTA) equipped with Dropout, including T2T14, T2T19, VOLOd1 and VOLOd2 in Table~\ref{tab:imgnet}.
We have these findings: (1) Clearly, our DropKey obtains significant enhancement against pure backbone.
Meanwhile, We note that Dropout can only bring a little boost to backbones and it even would degrade the performance of the model in some cases (eg. T2T14).
This indicates that the adaptive smoothing operator is more helpful to encourage the model to capture general information in large-scale datasets.
(2) Compare to lightweight models, heavyweight often need the larger drop ratio to achieve a remarkable boost.
For T2T14 and VOLOd1, the best accuracy is obtained under the dr=0.05.
Instead, T2T19 and VOLOd2 achieve the best accuracy when dr=0.1, which suggests heavyweight models adopt a larger drop ratio to prevent overfitting.

\begin{table}
  \caption{ Comparisons on backbones with Dropout or DropKey on ImageNet. dr denotes drop ratio.
  }
  \label{tab:imgnet}
  \centering
  \scriptsize
  \begin{tabular}{cc|cc}
    \toprule
     Model & Top-1 & Model & Top-1 \\
    \cmidrule(r){1-4}
    T2T14~\citep{t2t} & 81.76 $\pm$ 0.05 & VOLOd1~\citep{volo} & 84.27 $\pm$ 0.12 \\
    T2T14~\citep{t2t} + Dropout(dr=0.05) & 81.65 $\pm$ 0.04 &VOLOd1~\citep{volo} +  Dropout(dr=0.05) & 84.35 $\pm$ 0.11 \\
    T2T14~\citep{t2t} + Dropout(dr=0.1) & 81.53 $\pm$ 0.08 &VOLOd1~\citep{volo} +  Dropout(dr=0.1) & 84.31 $\pm$ 0.04 \\
    T2T14~\citep{t2t} + DropKey(dr=0.05) & \textbf{82.04 $\pm$ 0.14} &VOLOd1~\citep{volo}+ DropKey(dr=0.05) & \textbf{84.53 $\pm$ 0.03} \\
    T2T14~\citep{t2t} + DropKey(dr=0.1) &  81.93 $\pm$ 0.09 &VOLOd1~\citep{volo} +  DropKey(dr=0.1) &  84.39 $\pm$ 0.02\\
    \cmidrule(r){1-4}
    T2T19~\citep{t2t} & 82.56 $\pm$ 0.09 & VOLOd2~\citep{volo} & 85.24 $\pm$ 0.05\\
    T2T19\citep{t2t} + Dropout(dr=0.05) & 82.63 $\pm$ 0.05 &VOLOd2~\citep{volo} +  Dropout(dr=0.05) & 85.22 $\pm$ 0.08\\
    T2T19\citep{t2t} + Dropout(dr=0.1) & 82.68 $\pm$ 0.04 &VOLOd2~\citep{volo} +  Dropout(dr=0.1) & 85.21 $\pm$ 0.03\\
    T2T19\citep{t2t} + DropKey(dr=0.05) & 82.71 $\pm$ 0.02 &VOLOd2~\citep{volo} +  DropKey(dr=0.05) & 85.33 $\pm$ 0.12\\
    T2T19\citep{t2t} + DropKey(dr=0.1) & \textbf{82.94 $\pm$ 0.04} &VOLOd2~\citep{volo} + DropKey(dr=0.1) & \textbf{85.38 $\pm$ 0.06}\\
    \bottomrule
  \end{tabular}
\end{table}

\begin{figure}[t!]
	\centering
	\begin{minipage}{0.56\linewidth}
		\centering
		\begin{table}[H]
          \caption{ Comparison of DETR with Dropout and DropKey on COCO validation set.
          }
          \label{tab:coco}
          \centering
          \tiny
          \begin{tabular}{c|cccccc}
            \toprule
             Model & AP& AP$_{50}$ & AP$_{75}$ & AP$_{S}$ & AP$_{M}$ & AP$_{L}$\\
             \cmidrule(r){1-7}
             DETR~\citep{detr} & 41.9 & 62.1 & 44.2 & 20.6 & 45.7 & 60.7\\
             DETR~\citep{detr} + Dropout & 42.2 & 62.3 & 44.3 & 20.8 & 45.9 & 61.1\\
             DETR~\citep{detr} + DropKey & \textbf{42.9} &\textbf{63.4} & \textbf{44.7 } & \textbf{21.1} & \textbf{46.7} & \textbf{61.8}\\
            \bottomrule
          \end{tabular}
        \end{table}
	\end{minipage}
	\
	\begin{minipage}{0.4\linewidth}
		\centering
		\begin{table}[H]
          \caption{ Comparison of METRO with Dropout and DropKey on HUMBI. 
          }
          \label{tab:humbi}
          \centering
          \tiny
          \begin{tabular}{c|ccc}
            \toprule
             Model & mPVE$\downarrow$ & mPJPE$\downarrow$ & PA-mPJPE$\downarrow$\\
             \cmidrule(r){1-4}
             METRO~\citep{metro} & 57.8& 51.5 & 39.9\\
             METRO~\citep{metro} + Dropout & 57.9 & 51.6& 39.7\\
             METRO~\citep{metro} + DropKey & \textbf{57.1} & \textbf{51.0} & \textbf{39.5}\\
            \bottomrule
          \end{tabular}
        \end{table}
        
	\end{minipage}
\end{figure}
\subsection{Object Detection in COCO}
In this section, we verify the generalizability and effectiveness of DropKey for object detection on COCO dataset~\citep{coco}.
We apply DETR~\citep{detr} framework for the experiments.
Specifically, DETR is composed of the backbone, encoder and decoder. 
We followed the model architecture, anchor definition and training recipe in ~\citep{detr} to build DETR+Dropout and DETR+DropKey.
In Table~\ref{tab:coco}, we report the results of DETR~\cite{detr} with Dropout or DropKey in terms of AP.
It can be seen that, our DropKey significantly outperforms existing Dropout.
In detail, DropKey achieve +0.7, +1.1, +0.4, +0.3, +0.8 and +0.7 higher AP than the Dropout, which is regarded as a remarkable boost considering the challenge on this benchmark.
Another observation is that large objects benefit more from DropKey than small objects.
These results suggest that DropKey has the advantage over Dropout.

\subsection{Human-Object Interaction Detection in HICO-DET}
Since the task of scene graph and relation understanding is quite sensitive to global context learning, so we also verify the effectiveness of DropKey in QPIC~\citep{qpic} for Human-Object Interaction Detection in HICO-DET~\citep{hico}. 
QPIC is a transformer-based feature extractor that can effectively aggregate contextually important information.
We respectively introduce Dropout and DropKey to both encoder and decoder in QPIC and train these models with the same training recipe as ~\citep{qpic}. 
The model named "QPIC-ResNet101 + Dropout (dr=0.1)" is the result of our reproduction of QPIC with $Scheduled\downarrow$ drop ratio (Official realization is equipped with $Constant$ drop ratio). The experimental results in Table~\ref{tab:hico}suggest DropKey has higher performance gain on the different evaluating indicators which are consistent with our expectations.

\begin{table}
          \caption{ Comparison of QPIC with Dropout and DropKey on HICO-DET.
          }
          \label{tab:hico}
          \centering
          \tiny
          \begin{tabular}{c|cccccc}
            \toprule
             Model & \makecell[c]{full\\ (Default)}& \makecell[c]{rare\\ (Default)} & \makecell[c]{non-rare\\(Default)} & \makecell[c]{full\\(Known object)} & \makecell[c]{rare \\(Known object) }& \makecell[c]{non-rare\\(Known object)}\\
             \cmidrule(r){1-7}
             QPIC-ResNet101~\citep{qpic} + Dropout (dr=0.1) & 29.96 & 24.03 & 31.63 & 32.42 & 26.01 & 34.31\\
             QPIC-ResNet101~\citep{qpic} + Dropout (dr=0.3) & 30.02 & 24.17 & 31.72 & 32.49 & 26.13 & 34.41\\
             QPIC-ResNet101~\citep{qpic} + DropKey (dr=0.3) & \textbf{30.87} &\textbf{24.63} & \textbf{32.24 } & \textbf{33.21} & \textbf{26.66} & \textbf{34.94}\\
            \bottomrule
          \end{tabular}
        \end{table}

\subsection{Human Body Mesh Recovery in HUMBI}
We verify the efficacy of DropKey for human body mesh task on HUMBI~\citep{humbi}.
We follow conventions to train with 294 subjects from scratch and test with 120 subjects. We use Mean Per Vetex Error (MPVE)~\citep{mpve}, Mean Per Joint Position Error (MPJPE)~\citep{mpjpe} and Procrustes Analysis MPJPE (PA-MPJPE)~\citep{pampjpe} as metrics.
We use public METRO~\citep{metro} as the baseline to apply DropKey to self-attention layer.
Table~\ref{tab:humbi} summaries the results. we can see that DropKey consistently improves the performance for all standard metrics.
Specifically, DropKey brings about 0.7, 0.5 and 0.4 on mPVE, mPJPE and PA-mPJPE, respectively, while Dropout only boosts 0.2 performance on one metric. These results further verify the generalizability of DropKey as well as its superiority over vanilla dropout.

\begin{figure}
  \centering
  \includegraphics[width=\textwidth]{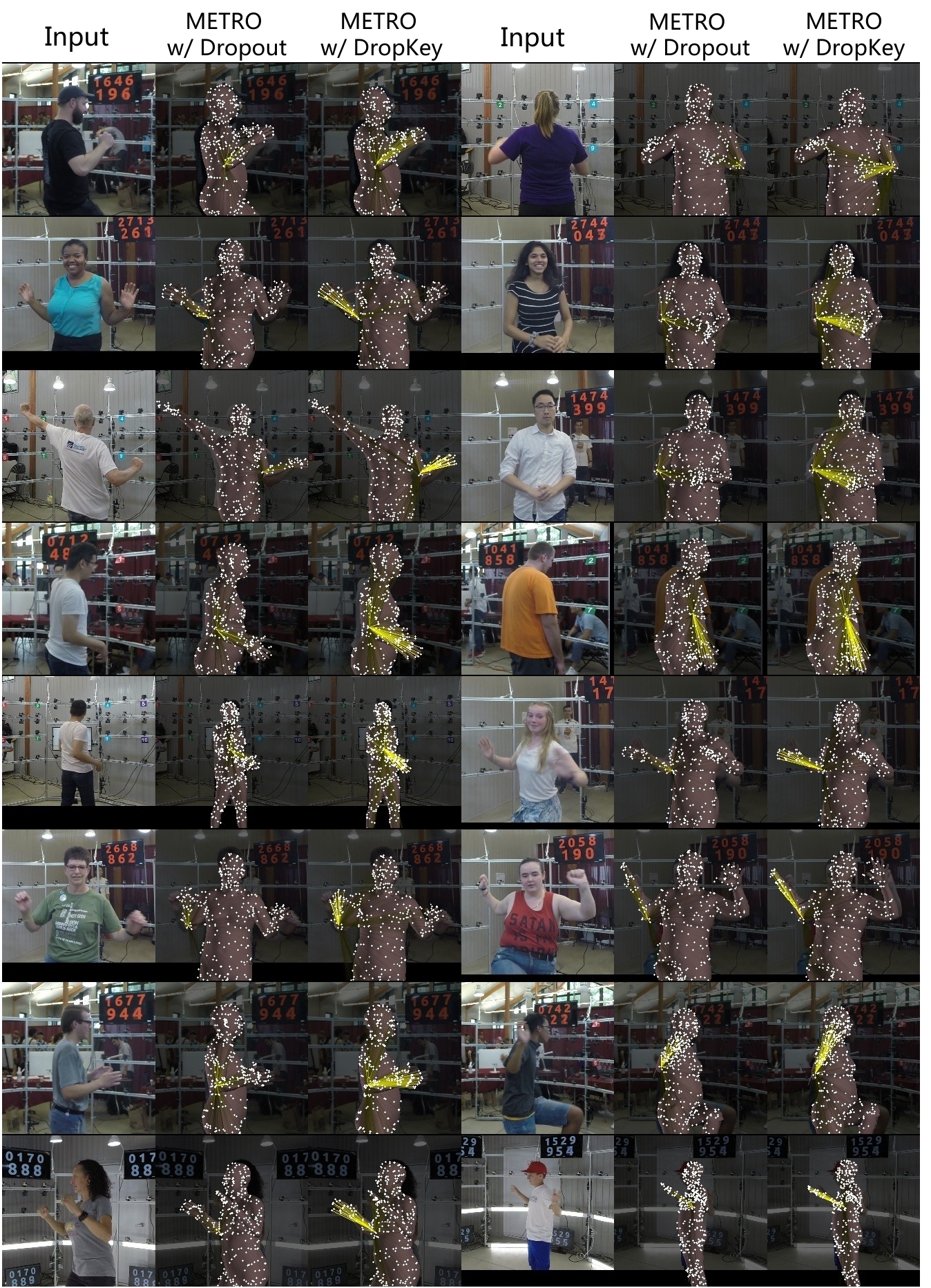}
  \caption{DropKey encourages model to learn robust interactions among body joints and mesh vertices for human mesh reconstruction. 
  Given an input image, METRO w/ Dropout predicts elbow joint only by taking sparse interactions with mesh vertices which is related to elbow  into consideration.
  DropKey encourages model to capture dense interactions with highly correlated vertices to learn the robust representation.}
 \label{fig:humbi}
\end{figure}

We also present visualization on HUMBI in Figure~\ref{fig:humbi} and results show that DropKey can encourage the model to capture dense interactions with highly correlated vertices to learn the robust representation.
METRO w/ Dropout only focuses on vertices around elbow joint which ignores global context.
For METRO w/ DropKey, it considers the interactions with vertices (eg. wrist joint and arm) which are helpful to predict the precise location of target joints.
This further demonstrates that the proposed DropKey can stimulate the model to capture vital information in a global manner.

\section{Conclusion}
In this paper, we explore the drop unit, drop schedule and structured necessity of the dropout technique in ViT. Specifically, we propose to set Key as the drop unit, which yields a novel dropout-before-softmax scheme. We theoretically and experimentally verify that this scheme can regularize attention weights and meanwhile and keep them as a probability  distribution simultaneously, helping capture vital patterns in a global manner and overcome local-bias problems that occurred to vanilla dropout. In addition, we present a new decreasing schedule for drop ratio, which stabilizes the training phase by avoiding overfittings in low-level features and maintaining sufficient high-level features. Moreover, we also experimentally show that structured dropout is not necessary for ViT. We distill the above analysis as a novel DropKey method, which plays as an improved version of dropout for ViT. Comprehensive experiments with different architectures on various vision tasks demonstrate the effectiveness of the proposed DropKey for enhancing ViTs.

\bibliographystyle{plain} 
\bibliography{neurips_2022}

\end{document}